\documentclass{article}

\usepackage[nonatbib,preprint]{neurips_2019}

\usepackage[utf8]{inputenc} 
\usepackage[T1]{fontenc}    
\usepackage{hyperref}       
\usepackage{url}            
\usepackage{booktabs}       
\usepackage{amsfonts}       
\usepackage{nicefrac}       
\usepackage{microtype}      

\usepackage[pdftex]{graphicx}
\usepackage{tikz}

\title{Impact of novel aggregation methods for flexible, time-sensitive EHR prediction without variable selection or cleaning}

\author{%
  Jacob Deasy, Ari Ercole and Pietro Li\`{o}\\
  University of Cambridge\\
  \texttt{\{jd645, ae105, pl219\}@cam.ac.uk}
}

\begin{document}

\maketitle

\begin{abstract}
    Dynamic assessment of patient status (e.g. by an automated, continuously updated assessment of outcome) in the Intensive Care Unit (ICU) is of paramount importance for \emph{early alerting, decision support} and \emph{resource allocation}. Extraction and cleaning of expert-selected clinical variables discards information and protracts collaborative efforts to introduce machine learning in medicine. We present improved aggregation methods for a \emph{flexible deep learning architecture} which learns a joint representation of patient chart, lab and output events. Our models outperform recent deep learning models for patient mortality classification using ICU timeseries, by embedding and aggregating all events with \emph{no pre-processing or variable selection}. Our model achieves a strong performance of AUROC 0.87 at 48 hours on the MIMIC-III dataset while using 13,233 unique un-preprocessed variables in an interpretable manner via hourly softmax aggregation. This demonstrates how our method can be easily combined with existing electronic health record systems for automated, dynamic patient risk analysis.
\end{abstract}

\section{Introduction}

In the United States, over a third of hospitals now utilize Electronic Health Record (EHR) databases that are considered broad enough to be comprehensive \cite{adler2015electronic}. The comprehensive, but complex, information contained in such EHRs gives a full description of the patient's clinical journey and is a promising source of data not only for retrospective studies, but also for building sophisticated decision support systems to improve care delivery \cite{johnson2016machine}.

However, a systematic review of 107 predictive models built with EHR data found that only 34.6\% (37/107) used longitudinal data, and the median number of variables used was only 27 \cite{goldstein2017opportunities,rajkomar2018scalable}. The Intensive Care Unit (ICU) has particularly high longitudinal data density since patients' conditions may change on timescales of minutes. Despite this, current mortality risk estimates are often solely based on acute physiology scores \cite{desai2019}. These metrics are static and typically reliant upon logistic regression of specific markers of patient physiology recorded during the first hours after ICU admission. Such approaches are likely to both be sub-optimal in terms of predictive power and unsuitable for decision support as they cannot be updated in real-time.

This work introduces novel aggregation methods for a flexible method of learning from \emph{all chart, lab and output events}, regardless of type, frequency or cardinality. Our deep learning architecture achieves strong in-hospital mortality prediction at 48 hours of AUROC 0.878 (95\% CI: 0.871-0.883) on patients in the MIMIC-III dataset \cite{johnson2016mimic}, without performing any variable selection or pre-processing. Instead, each model learns a powerful method of embedding and aggregating discrete variable categories from raw EHR data. We assess an aggregation variant which builds in model interpretability and find that the predictive performance cost is negligible---with models still performing in line with recent strong deep learning architectures. Any sequential EHR data can be incorporated into our model to provide more interpretable and personalized dynamic risk prediction, with reliable dynamic performance exceeding traditional severity scores after only a few hours in the ICU.

\section{Related Work}
\label{gen_inst}

Recurrent Neural Networks (RNNs) \cite{elman1990finding,rumelhart1988learning} have been shown to outperform traditional heuristic ICU scores for prediction of mortality and long Length Of Stay (LOS) risk \cite{johnson2016machine,rajkomar2018scalable}. Meanwhile, state-of-the-art models using sequential data have grown in size and complexity and been adapted to address the unique challenges of the ICU \cite{rajkomar2018scalable,che2018recurrent}. In all of these cases, the dominant approach to deep learning with EHR data remains reliant on an initial stage of variable selection involving the use of expert knowledge to hand-pick a subset of clinically relevant variables \cite{che2018recurrent,harutyunyan2017multitask}. Even on the broad MIMIC-III dataset, clinical variable selection has recently been somewhat standardized by the introduction of a benchmark \cite{harutyunyan2017multitask}. Larger studies have managed to avoid variable selection by building pipelines which convert their standard data type into a format suitable for machine learning \cite{rajkomar2018scalable,tabak2013using}. Although this has led to strong prediction results, these architectures do not generalize well to EHRs and regions where little effort is being made to produce a standardized EHR database.\\

On the other hand, previous research has shown that using sparsity-inducing priors in deep Bayesian neural networks can automatically reduce the number of clinical variables attended to by an \emph{AI clinician} \cite{komorowski2018artificial}, with little impact on performance \cite{popkes2019interpretable}---calling into question the need for variable selection in the first place. This has lead to the recent introduction of the first deep learning pipeline which only requires that EHR time series contain a sequence of timestamps and a sequence of variable readings \cite{deasy2019time}. We extend this work by exploring embedding aggregation variants that can process arbitrarily many patient readings per hour. We create a pipeline and dynamic model that perform zero variable selection, data processing or model ensembling with performance boosting aggregation variants. Our models aggregate varied patient demographic and physiological data in an interpretable manner to provide a more clinically-relevant and accurate outcome prediction method in the ICU. As well as exploring the power of embedding aggregation functions, by producing a time-sensitive model, our approach tackles the inherent and oft-neglected need for dynamic assessment in the ICU \cite{meiring2018optimal}.

\section{Model and inputs}
\label{model_and_inputs}

\subsection{Dataset}
\label{dataset}

\begin{table}[b]
    \caption{Mortality class distribution and ICU lengths of stay information for our  MIMIC-III subset.}
    \label{tab:data}
    \centering
    \begin{tabular}{llll}
        \toprule
        Model          & Patients        & Mean LOS      & Median LOS \\
        \midrule
        Died           & 2,797 (13.2\%)  & 8.11        & 5.35 \\
        Survived       & 18,342 (86.8\%) & 5.65        & 3.50 \\
        \midrule
        \textbf{Total} & \textbf{21,139} & \textbf{5.97} & \textbf{3.72} \\
        \bottomrule
    \end{tabular}
\end{table}

We used three types of data from the MIMIC-III dataset for classification: (a) \emph{chart event} recordings of routine vital signs from the electronic chart forming the bulk of patient information, (b) \emph{lab event} results of laboratory tests on blood and urine, and (c) \emph{output event} recordings of admission information and treatments. The objective is to dynamically assign a risk of mortality to each patient based on their event time series during the first 48 hours after their admission to the ICU.

The MIMIC-III database contains high-resolution patient data, including: demographics, vital sign time-series, laboratory tests, illness severity scores, medications and procedures, fluid intake and outputs, clinician notes, and diagnostic coding.  The median age of adult patients is 65.8 years ($Q_1$–$Q_3$: 52.8–77.8), 55.9\% of patients are male, and in-hospital mortality is 11.5\% \cite{johnson2016mimic}. Between the two EHR systems that comprise MIMIC-III, CareVue and MetaVision, in total the overall dataset contains 330,712,483 chart events, 27,854,055 lab events and 4,349,218 output events.

The architecture was trained using a subset of patient stays where the proportion of in-hospital deaths was 13.2\%. The proportion of long ICU stays (greater than 7 days in the ICU) was 23.0\% (4,868/21,139)---see Table~\ref{tab:data} for a summary of mortality and LOS statistics.

\subsection{Data pipeline}
\label{data_pre-processing}

Unlike traditional approaches, we retain all of the chart, lab and output events for each stay without any data cleaning, outlier removal or domain-specific knowledge. The processing we perform is enough to assign a patient ID, a stay ID and a timestamp to each event---a pipeline which is independent of EHR data formatting or structure. Our model takes the entire patient timeseries as input, regardless of event type, frequency or cardinality and discretizes each event (see Table~\ref{tab:labels}). We note that, because we do not select for clinical variables, after event association with patient stays, our EHR dataset contains 208,572,237 events instead of the 31,868,114 employed in \cite{harutyunyan2017multitask} and all subsequent papers relying on the MIMIC-III benchmark. Due to the increased number of variables used by our model, after processing we also have a higher number of patients and stays available, supporting the evidence in favour of models which can incorporate broad EHR data.

\begin{table}
    \caption{Examples of discrete token creation alongside percentile-based quantization and tokenization of continuous variable. Discrete variables remain discrete, while values that can be converted to floating point numbers are considered continuous and separated into 20 percentile-based bins.}
    \label{tab:labels}
    \centering
    \begin{tabular}{llll}
        \toprule
        Label & Value & Percentile & Token \\
        \midrule
        Blood pH & 7.41 & 55-60\% & Blood pH\_12 \\
        Heart Rate & 69 & 35-40\% & Heart Rate\_8 \\
        Code Status & Full Code &  & Code Status Full Code \\
        \bottomrule
    \end{tabular}
\end{table}

\subsection{Model architecture}
\label{model_architecture}

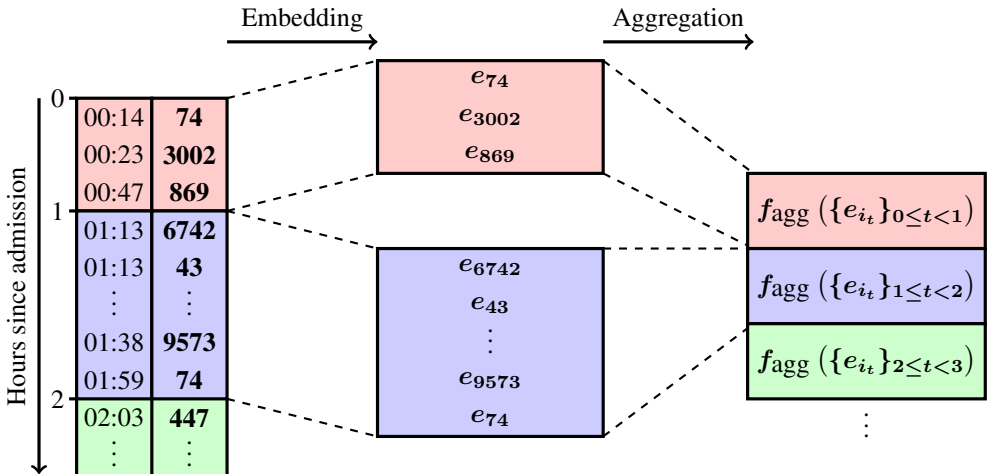
\begin{figure}[h]
    \centering
    \usetikzlibrary{positioning}

\begin{tikzpicture}

\node at (-0.25, 4.0) {0};
\draw[very thick] (-0.1, 4.0) -- (0.0, 4.0);
\node at (-0.25, 2.5) {1};
\draw[very thick] (-0.1, 2.5) -- (0.0, 2.5);
\node at (-0.25, 0.0) {2};
\draw[very thick] (-0.1, 0.0) -- (0.0, 0.0);
\draw[->,very thick] (-0.5, 4) -- (-0.5, -1);
\node[fill=white,rotate=90] at (-0.8, 1.5) {Hours since admission};

\node[draw,rectangle,very thick,minimum width=1cm,minimum height=1.5cm,fill=red!20] at (0.5, 3.25) {};
\node[draw,rectangle,very thick,minimum width=1cm,minimum height=1.5cm,fill=red!20] at (1.5, 3.25) {};
\node[rectangle,minimum width=1cm,minimum height=1cm,fill=green!20] at (0.5, -0.5) {};
\node[rectangle,minimum width=1cm,minimum height=1cm,fill=green!20] at (1.5, -0.5) {};
\node[draw,rectangle,very thick,minimum width=1cm,minimum height=2.5cm,fill=blue!20] at (0.5, 1.25) {};
\node[draw,rectangle,very thick,minimum width=1cm,minimum height=2.5cm,fill=blue!20] at (1.5, 1.25) {};
\draw[very thick] (0, 0) -- (0, -1);
\draw[very thick] (1, 0) -- (1, -1);
\draw[very thick] (2, 0) -- (2, -1);

\node at (0.5, 3.75)    {00:14};
\node at (0.5, 3.25)    {00:23};
\node at (0.5, 2.75)    {00:47};
\node at (0.5, 2.25)    {01:13};
\node at (0.5, 1.75)    {01:13};
\node at (0.5, 1.375)   {$\vdots$};
\node at (0.5, 0.75)    {01:38};
\node at (0.5, 0.25)    {01:59};
\node at (0.5, -0.25)   {02:03};
\node at (0.5, -0.625) {$\vdots$};

\node at (1.5, 3.75) {\textbf{74}};
\node at (1.5, 3.25) {\textbf{3002}};
\node at (1.5, 2.75) {\textbf{869}};
\node at (1.5, 2.25) {\textbf{6742}};
\node at (1.5, 1.75) {\textbf{43}};
\node at (1.5, 1.375) {$\vdots$};
\node at (1.5, 0.75) {\textbf{9573}};
\node at (1.5, 0.25) {\textbf{74}};
\node at (1.5, -0.25) {\textbf{447}};
\node at (1.5, -0.625) {$\vdots$};

\draw[thick,dashed] (2, 0.0) -- (4, -0.5);
\draw[thick,dashed] (2, 2.5) -- (4, 2.0);
\node[draw,rectangle,very thick,minimum width=3cm,minimum height=2.5cm,fill=blue!20] at (5.5, 0.75) {};
\node[font=\boldmath] at (5.5, 1.75)  {$e_{6742}$};
\node[font=\boldmath] at (5.5, 1.25)  {$e_{43}$};
\node[font=\boldmath] at (5.5, 0.875) {$\vdots$};
\node[font=\boldmath] at (5.5, 0.25)  {$e_{9573}$};
\node[font=\boldmath] at (5.5, -0.25) {$e_{74}$};

\draw[thick,dashed] (2, 2.5) -- (4, 3);
\draw[thick,dashed] (2, 4.0) -- (4, 4.5);
\node[draw,rectangle,very thick,minimum width=3cm,minimum height=1.5cm,fill=red!20] at (5.5, 3.75) {};
\node[font=\boldmath] at (5.5, 4.25)  {$e_{74}$};
\node[font=\boldmath] at (5.5, 3.75)  {$e_{3002}$};
\node[font=\boldmath] at (5.5, 3.25)  {$e_{869}$};

\draw[->,very thick] (2, 4.75) -- node[above] {Embedding} (4, 4.75);
\draw[->,very thick] (7, 4.75) -- node[above] {Aggregation} (9, 4.75);

\draw[thick,dashed] (7, 4.5)  -- (9,3);
\draw[thick,dashed] (7, 3.0)  -- (9,2);
\draw[thick,dashed] (7, 2.0)  -- (9,2);
\draw[thick,dashed] (7, -0.5) -- (9,1);

\node[draw,rectangle,very thick,minimum width=3cm,minimum height=1cm,fill=red!20,font=\boldmath] at (10.5, 2.5) {$f_{\textrm{agg}}\left(\{e_{i_{t}}\}_{0\leq t<1}\right)$};
\node[draw,rectangle,very thick,minimum width=3cm,minimum height=1cm,fill=blue!20,font=\boldmath] at (10.5, 1.5) {$f_{\textrm{agg}}\left(\{e_{i_{t}}\}_{1\leq t<2}\right)$};
\node[draw,rectangle,very thick,minimum width=3cm,minimum height=1cm,fill=green!20,font=\boldmath] at (10.5, 0.5) {$f_{\textrm{agg}}\left(\{e_{i_{t}}\}_{2\leq t<3}\right)$};
\node at (10.5,-0.25) {$\vdots$};

\end{tikzpicture}
    \caption{Example of our flexible EHR embedding for a single patient time series. EHR token indices are embedded and reduced to a fixed size hourly representation by aggregation function $f_{\textrm{agg}}$.}
    \label{fig:tikz_diagram}
\end{figure}

We use an embedding layer followed by an aggregation function to reduce an arbitrary number of inputs to a fixed size representation (see Figure~\ref{fig:tikz_diagram}), before passing this to a recurrent neural network to classify patient mortality. All chart, lab and output events are tokenized and embedded; the resulting set of vectors are aggregated over each hour by one of the following: summation, average, weighted average or a softmax (masked by hour). Dynamic classification at each time step is achieved by a dense layer with sigmoid activation applied to each hidden state.

\section{Experiments and results}
\label{Experiments}

\subsection{Evaluation procedure}

We used $k$-fold cross-validation with $k = 10$ to assess model performance on a held out test set of 2,114 patients (10,000 bootstrap samples were used to construct confidence intervals). All models were trained for 50 epochs with an early stopping threshold of 5 epochs with no increase in AUROC on the validation set of 1,000 patients. All models were trained with a batch size of 128, using the Adam optimizer \cite{kingma2014adam}, with a learning rate of 0.001 and default hyperparameters otherwise. As the distribution of mortality across the MIMIC-III dataset is imbalanced (see Table~\ref{tab:data}), accuracy is an unsuitable metric so we employ AUROC score to evaluate model performance.

\subsection{Model parameters}

Our baseline model contains 38,780 embeddings in a 32D space, generated by discretizing 13,233 unique variable labels, with 20 bins per continuous variable. We utilize a GRU \cite{cho2014learning} of depth 1 with a sigmoid output activation, layer normalization applied to all linear projections \cite{lei2016layer}, dropout regularization \cite{srivastava2014dropout} ($p=0.5$) on both the aggregated embedding and the hidden states, as well as a weight decay coefficient of 0.001 \cite{krogh1992simple}. We assess the effect of using summation, simple averaging and a masked softmax (over a separate token weight vector) as aggregation functions on model performance. We note that the risk of unnormalized aggregation, such as summation, is that the model learns to count the number of readings taken in an hour and correlate this with patient outcome.

\subsection{Results}

Table~\ref{tab:results} demonstrates that integrating events in a unified manner results in strong early classification of in-hospital mortality (AUROC $0.88$)---stronger than both traditional ICU scores and recent deep learning models despite no feature engineering. AUROC of $0.88$ means that there is an $88$\% chance of assigning higher risk to a random patient destined to die than a random patient destined to live. If clinical resource allocation were based on our model rather than SAPS II, 25\% more patients would be correctly prioritized. In particular, the softmax aggregation model variant automatically learns comparative hourly weights for each EHR token, building in interpretability with negligible cost to performance. Exploratory work on this weighting system has validated the association of well-known mortality indicators, such as high respiratory rate, resuscitation instructions and even documentation of visits by the priest, and opens avenues for future research on patient-specific risk variation.

As such, this research presents a model that is flexible enough to reduce the requirements on sequential EHR data to simply timestamps alongside variable readings. Percentile-based quantization followed by embedding aggregation also means our model automatically learns how \emph{outliers} and \emph{missing values} influence patient mortality risk. By eliminating the need for variable selection and generating comparative weights while outperforming a strong deep learning baseline, our aggregation improvements allow machine learning specialists to consult clinicians about highly weighted variables \emph{after} model development, rather than before. Future work will focus on building models that learn to convert any of the diverse range of design principles and computational practices currently employed across EHR databases to the form used by our model.

\begin{table}
  \caption{Cross-validation results for prediction of mortality at 12 and 48 hours for the embedding aggregation functions, compared to models that make use of carefully selected predictor variables.}
  \label{tab:results}
  \centering
  \begin{tabular}{llll}
    \toprule
    Model/Aggregation    & \# curated features & 12h AUROC (95\% CI) & 48h AUROC (95\% CI) \\
    \midrule
    OASIS \cite{johnson2013new} & 10 & --- & 0.663 \\
    SAPS II \cite{le1993new} & 17 & --- & 0.705 \\
    Benchmark \cite{harutyunyan2017multitask} & 76 & --- & 0.870 (0.852-0.887) \\
    \midrule
    Weighted average \cite{deasy2019time} & --- & 0.798 (0.794-0.802) & 0.856 (0.851-0.861)  \\
    Masked softmax & --- & 0.805 (0.796-0.814) & 0.872 (0.862-0.881)  \\
    Average & --- & \textbf{0.811 (0.807-0.815)} & 0.875 (0.873-0.877)  \\
    Summation & --- & 0.810 (0.804-0.815) & \textbf{0.878 (0.871-0.884)}  \\
    \bottomrule
  \end{tabular}
\end{table}

\small

\bibliographystyle{vancouver}
\bibliography{ms}

\end{document}